\documentclass{article}
\usepackage{iclr2027,times}

\usepackage{amsmath,amsfonts,bm}

\def\eqref#1{equation~\ref{#1}}

\def\1{\bm{1}}

\DeclareMathAlphabet{\mathsfit}{\encodingdefault}{\sfdefault}{m}{sl}
\SetMathAlphabet{\mathsfit}{bold}{\encodingdefault}{\sfdefault}{bx}{n}

\usepackage{amsmath}
\usepackage{amssymb}
\usepackage{array}
\usepackage{caption}
\usepackage{enumitem}
\usepackage{graphicx}
\usepackage{booktabs}
\usepackage{multirow}
\usepackage{makecell}
\usepackage{float}
\usepackage{placeins}
\usepackage{wrapfig}
\usepackage{afterpage}
\usepackage{hyperref}
\hypersetup{hidelinks}
\usepackage{url}

\setlist[itemize]{leftmargin=*, itemsep=2pt, topsep=3pt}

\newcommand{\rcvlatableformat}{%
  \small
  \setlength{\tabcolsep}{4pt}%
  \renewcommand{\arraystretch}{1.08}%
}
\hypersetup{
  hidelinks,
  pdfauthor={},
  pdftitle={RCVLA: 4D Radar-Grounded Semantic Reasoning and Trajectory Arbitration for Autonomous Driving}
}

\title{RCVLA: 4D Radar-Grounded Semantic Reasoning and Trajectory Arbitration for Autonomous Driving}

\author{Lianqing Zheng\textsuperscript{1,*}, Xiaokai Bai\textsuperscript{2,*},
Yixuan Luo\textsuperscript{2}, Runwei Guan\textsuperscript{3},
Minghao Liu\textsuperscript{4}, \\[1mm]
\textbf{Zhiqiang Wei\textsuperscript{5}, Hui-liang Shen\textsuperscript{2},
Xichan Zhu\textsuperscript{1},
Zhixiong Ma\textsuperscript{1,$\dagger$}}\\[1mm]
\textsuperscript{1}Tongji,
\textsuperscript{2}ZJU,
\textsuperscript{3}HKUST,
\textsuperscript{4}UTokyo,
\textsuperscript{5}SJTU\\[1mm]
\textsuperscript{*}Equal Contribution. 
\textsuperscript{$\dagger$}Corresponding Author.\\[1mm]
TJRadarLab@163.com
}
\iclrfinalcopy
\begin{document}

\maketitle

\begin{abstract}
4D radar provides geometric and motion cues that complement visual semantics, 
but integrating it into vision-language-action (VLA) models requires both
radar--language alignment for semantic reasoning and explicit use of radar
measurements for trajectory refinement and selection. To support these
capabilities, we construct Cap4DR
with 86,016 radar-image-text samples for alignment pretraining and OmniHD-QA
with 520,161 question-answer pairs for instruction tuning across scene
description, key-object reasoning, occupancy understanding, and trajectory
planning. Building on these datasets, we propose RCVLA, a radar-camera VLA
framework consisting of a radar-grounded semantic
reasoning stage (RCVLA-Sem) and a trajectory arbitration stage
(RCVLA-Phys). RCVLA-Sem performs gated bidirectional interaction
between camera and radar tokens for driving question answering and reference
trajectory generation, while auxiliary heads provide object and occupancy
queries. RCVLA-Phys refines reference-guided trajectory candidates through truncated
diffusion conditioned on these queries and cluster-level radar measurements,
then calibrates candidate scores using radar-derived time-to-collision risk.
On OmniHD-QA, RCVLA-Sem improves CIDEr by 9.92 points and reduces key-object
velocity error by $21.9\%$ relative to OmniDrive. RCVLA-Phys further reduces 
average L2 error from $0.348$ to $0.259\,\mathrm{m}$ and
average open-loop collision rate from $0.576\%$ to $0.175\%$ relative to
RCVLA-Sem. Ablation studies further show that language-aligned radar tokens improve semantic reasoning,
while cluster-level radar measurements and risk calibration improve trajectory
arbitration. Code will be released.

\end{abstract}

\section{Introduction}
\label{sec:introduction}

Recent vision-language-action (VLA) models for autonomous driving 
couple language-based scene reasoning with driving decisions and motion planning
\citep{sima2024drivelm,wang2025omnidrive,jiang2026senna}. However, camera-centric 
VLA models infer geometry and motion primarily from visual observations, which can 
become unreliable under degraded visibility \citep{bijelic2020seeing}.
4D radar supplies spatial measurements and
Doppler-derived radial velocity that complement visual semantics
\citep{palffy2022multi,zhang2023ntu4dradlm,choi2023msc}.

Integrating 4D radar into VLA models raises two challenges. First, standard
image--text pretraining does not provide 4D radar--language alignment,
motivating dedicated alignment supervision and driving instruction data.
Second, semantic reasoning and trajectory planning require different 4D radar representations. 
Compact language-aligned tokens provide an interface for scene reasoning \citep{li2023blip}, while trajectory
refinement and selection rely on local geometric and motion measurements.
We therefore retain cluster-level radar measurements to guide candidate
refinement and risk-aware selection, which together constitute
\emph{trajectory arbitration}.

\begin{figure}[t]
  \centering
  \includegraphics[width=0.90\textwidth]{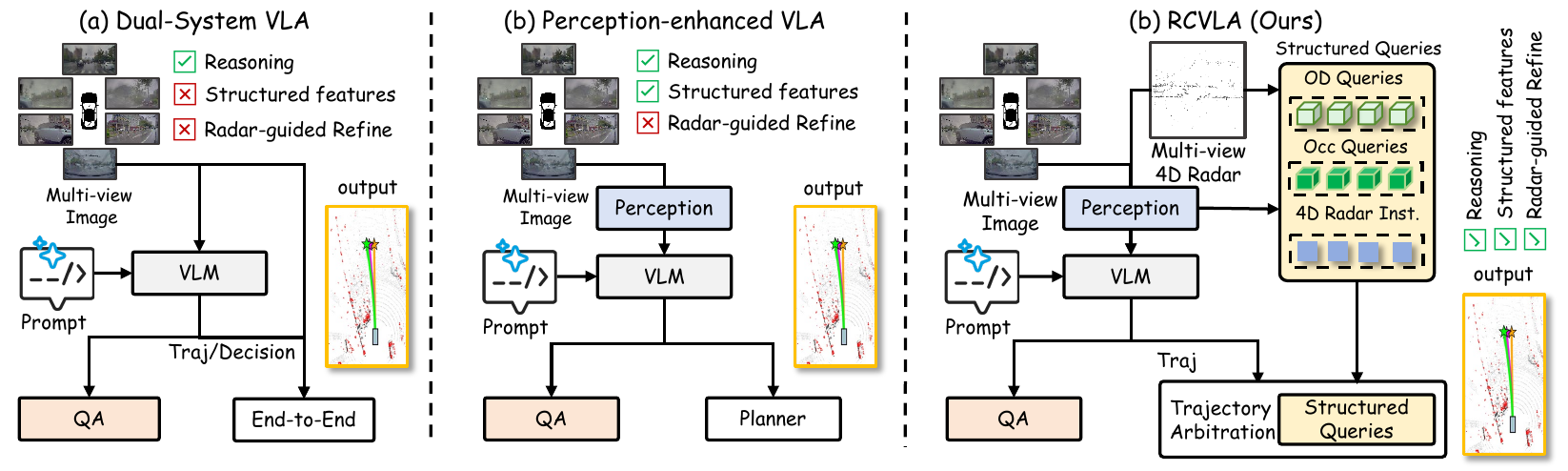}
  \caption{\textbf{Architectural comparison.} (a) Dual-system VLA links
  language reasoning to a separate end-to-end model. (b) Perception-enhanced VLA
  incorporates structured scene features. (c) RCVLA combines a semantic
  reference, structured scene queries, and 4D radar measurements for
  trajectory refinement and risk-aware selection.}
  \label{fig:rcvla_head}
\end{figure}

To address these challenges, we construct two datasets and
develop \textbf{RCVLA}. \textbf{Cap4DR} contains 86,016 4D radar-image-text
samples for radar--language alignment pretraining, with 4D radar relevant
entities verified through multi-model cross checking and selective manual
review. \textbf{OmniHD-QA} contains 520,161 question-answer pairs derived from
OmniHD-Scenes \citep{zheng2026omnihd}, covering scene description, key-object
reasoning, occupancy understanding, and trajectory planning for instruction
tuning and evaluation. Building on these datasets, RCVLA integrates
radar-grounded semantic reasoning with trajectory arbitration. As illustrated
in Figure~\ref{fig:rcvla_head}, we compare three VLA designs. Dual-system
VLA couples language reasoning with a separate end-to-end model, while
perception-enhanced VLA further incorporates structured scene features.
RCVLA instead combines a reference trajectory from semantic reasoning,
structured scene queries, and direct 4D radar measurements for trajectory
refinement and risk-aware selection.

\vspace{-0.4em}

\raggedbottom
Specifically, \textbf{RCVLA-Sem} performs gated bidirectional interaction
between camera and 4D radar tokens for driving question answering and reference
trajectory generation, while auxiliary object and occupancy heads provide
structured scene queries. \textbf{RCVLA-Phys} constructs a compact candidate
set from the reference trajectory and anchors, then refines it
through truncated diffusion conditioned on these queries and cluster-level
radar measurements. The retained measurements further yield
time-to-collision (TTC)-based risk estimates that calibrate candidate scores
for risk-aware selection. Our contributions are summarized below.

\begin{itemize}[leftmargin=10pt,itemsep=0.35\baselineskip,
  topsep=0.35\baselineskip,parsep=0pt,partopsep=0pt]

  \item We construct \textbf{Cap4DR} and \textbf{OmniHD-QA}, providing
  complementary supervision for radar--language alignment pretraining,
  multitask driving instruction tuning, and evaluation.

  \item We propose \textbf{RCVLA}, a radar-camera VLA framework that uses 4D
  radar in complementary roles for semantic reasoning and trajectory
  arbitration. RCVLA-Sem integrates camera and radar tokens for driving
  question answering and reference trajectory generation, while RCVLA-Phys
  refines trajectory candidates using structured scene queries and
  cluster-level radar measurements, with radar-derived TTC risk guiding
  candidate selection.

  \item Experiments on \textbf{OmniHD-QA} demonstrate improvements in semantic
  reasoning and open-loop planning over the evaluated baselines. Ablations
  further validate the complementary roles of language-aligned radar tokens
  in scene reasoning and preserved physical measurements in trajectory
  refinement and risk-aware selection.

\end{itemize}
\section{Related Work}
\label{sec:related_work}

\subsection{Planning-Oriented End-to-End Driving}
End-to-end driving predicts driving actions or future trajectories directly
from sensor observations within a unified framework. ST-P3 integrates perception, prediction, and
planning with a shared spatio-temporal representation \citep{hu2022st},
while UniAD coordinates multiple driving tasks through a unified query
interface \citep{hu2023planning}. BEV-Planner studies the role of ego-state information in open-loop 
planning \citep{li2024ego}, while DiffusionDrive uses trajectory anchors and truncated diffusion to 
generate diverse future trajectories \citep{liao2025diffusiondrive}. These methods connect structured scene 
perception with motion planning, but are mainly designed for predefined driving tasks and offer limited scene-level reasoning.

\subsection{Vision-Language-Action Models for Driving}
Vision-language models (VLMs) have shown strong capabilities in visual 
understanding and reasoning \citep{liu2023visual,chen2024internvl}.
Recent methods extend these capabilities to scene reasoning and motion
planning. DriveLM organizes perception, prediction, and planning through
graph-based question answering \citep{sima2024drivelm}. OmniDrive combines 
multi-view 3D scene understanding with trajectory generation \citep{wang2025omnidrive}, 
while Senna uses a large vision-language model to support decision making within an end-to-end driving 
framework \citep{jiang2026senna}. These methods bring language reasoning into autonomous driving, 
but still rely mainly on visual observations, so geometric and motion information is inferred rather than directly measured..

\subsection{4D Radar Perception and Language Grounding}
Recent studies connect 4D radar perception with language models. WaterVG and NanoMVG
combine language, camera, and 4D radar inputs for grounding in waterway
scenes \citep{guan2025watervg,guan2025nanomvg}. Talk2Radar grounds referring
expressions in 4D radar point clouds \citep{guan2025talk2radar}, while
Text4Radar-V2X introduces language guidance into cooperative 4D radar-based 3D
detection \citep{peng2026text4radar}. However, these studies mainly use 4D radar for 
grounding and detection, while its physical measurements have not been fully exploited for 
scene understanding and downstream trajectory planning.
\section{Datasets}
\label{sec:datasets}

Existing image--text datasets are largely centered on visual inputs, with
little consideration of 4D radar observations. We therefore construct Cap4DR for
radar--language pretraining and OmniHD-QA for multitask instruction tuning
and evaluation.

\begin{figure*}[t]
  \centering
  \includegraphics[height=0.163\textheight,keepaspectratio]
  {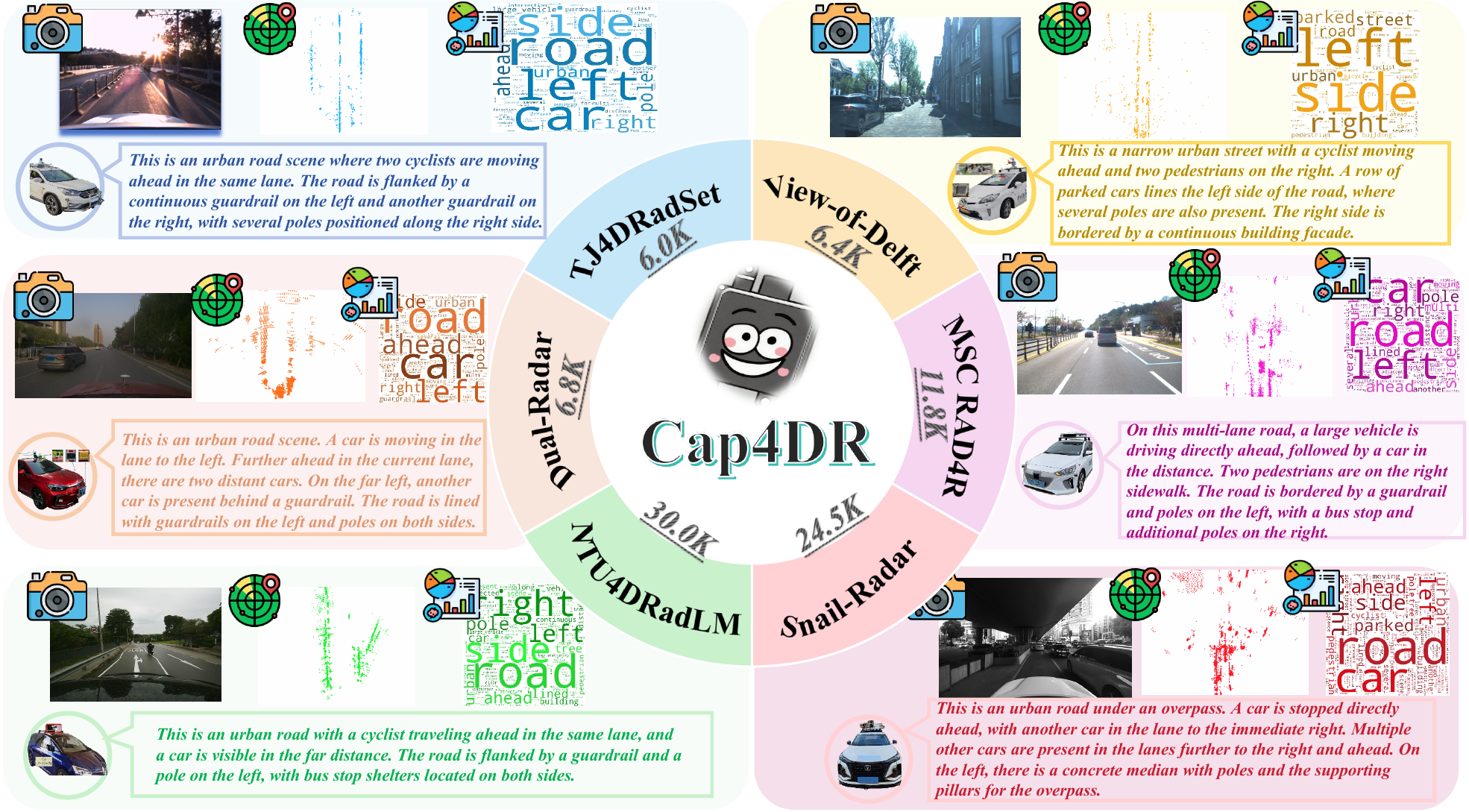}\hfill
  \includegraphics[height=0.163\textheight,keepaspectratio]
  {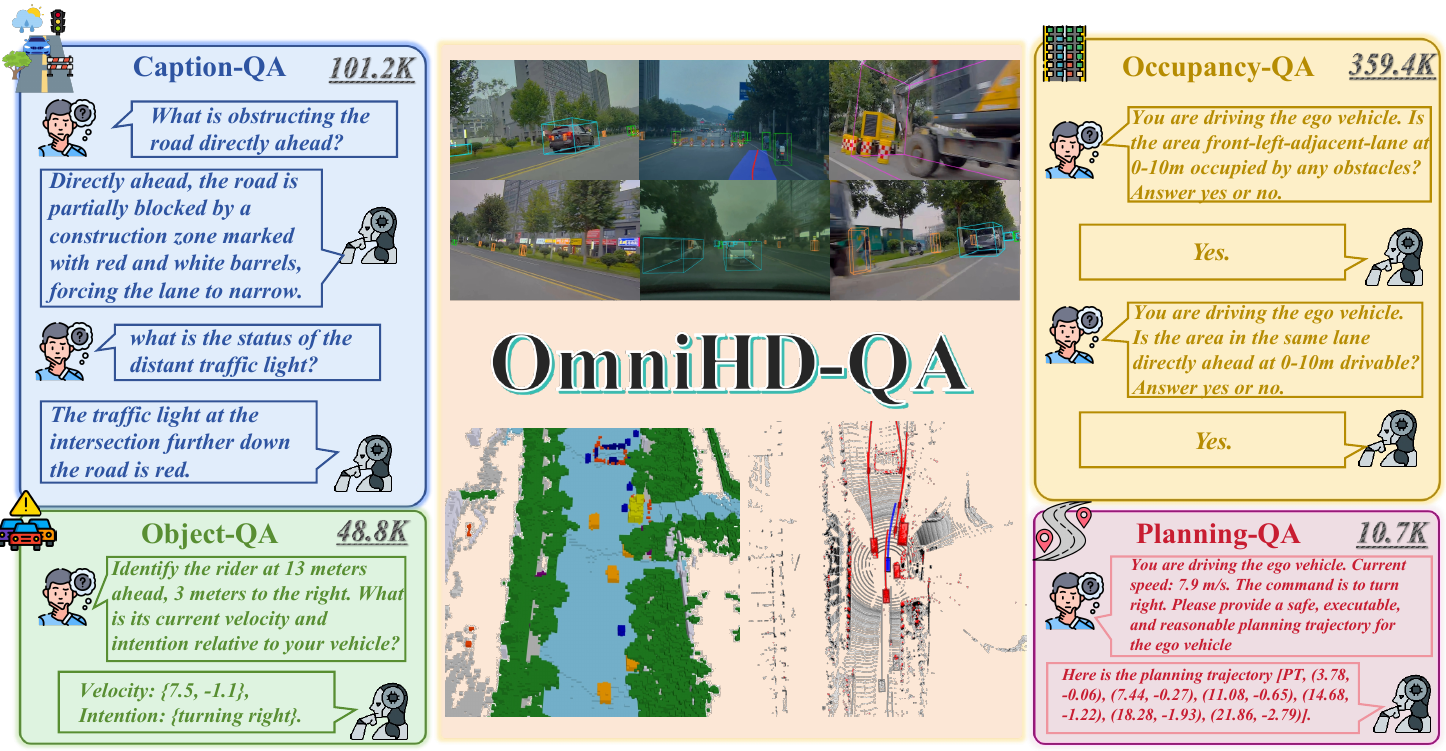}
  \caption{\textbf{Dataset overview.} Left: Cap4DR combines samples from six
  radar datasets with radar-oriented descriptions for alignment pretraining. Right:
  OmniHD-QA covers scene description, key-object reasoning, 3D occupancy,
  and trajectory planning for instruction tuning and evaluation.}
  \label{fig:dataset_overview}
\end{figure*}

\noindent\textbf{Cap4DR.} Cap4DR integrates synchronized front-view images and 
4D radar data from View-of-Delft \citep{palffy2022multi}, TJ4DRadSet \citep{zheng2022tj4dradset}, 
NTU4DRadLM \citep{zhang2023ntu4dradlm}, MSC-RAD4R \citep{choi2023msc}, SNAIL Radar \citep{huai2025snail}, 
and Dual Radar \citep{zhang2025dual}. We first apply temporal alignment and unified radar preprocessing 
across datasets. Rather than using generic image captions directly, we retain only objects and spatial 
relations that can be associated with radar observations, turning visual descriptions into radar-relevant 
language supervision. The extracted entities are cross-checked by three VLMs, with 
ambiguous cases manually reviewed. Our Cap4DR dataset contains 86,016 radar--text 
pairs for radar--language alignment pretraining.  

\noindent\textbf{OmniHD-QA.} OmniHD-QA is constructed from OmniHD-Scenes
\citep{zheng2026omnihd} and contains 520,161 question--answer pairs covering
scene description, key-object reasoning, occupancy understanding, and
trajectory planning. We use Gemini2.5-Pro \citep{comanici2025gemini} to generate
scene-description QA from multi-view images with available annotations. The remaining tasks are constructed from
structured annotations, including object tracks, semantic occupancy, 
and ego pose. OmniHD-QA therefore provides unified instruction supervision for
driving-specific scene reasoning and planning.
The reported datasets use the synchronization, annotation, and split rules
described in this paper.

\flushbottom
\section{Method}
\label{sec:method}
\subsection{Overview}
\label{sec:semantic_physical_spaces}
\begin{figure}[t]
  \centering
  \includegraphics[width=0.86\linewidth]{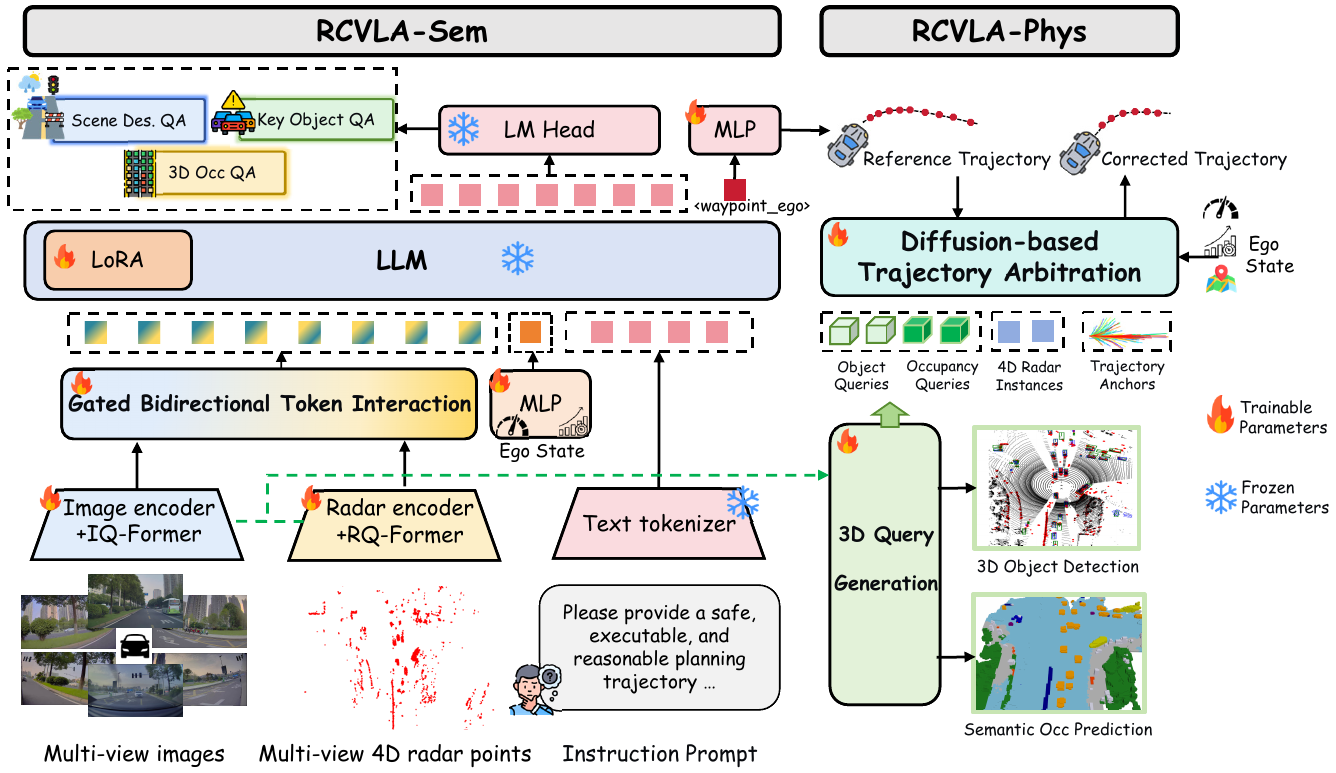}
  \caption{\textbf{RCVLA architecture.} RCVLA-Sem integrates camera and radar tokens for driving question answering
and long-horizon reference planning, while providing structured object and
occupancy queries. RCVLA-Phys combines the reference trajectory, structured
queries, and cluster-level radar measurements for trajectory refinement and
TTC-aware selection.}
  \label{fig:rcvla_overview}
\end{figure}

\begin{wrapfigure}[23]{r}{0.60\textwidth}
  \vspace{-\intextsep}
  \centering
  \includegraphics[width=\linewidth]{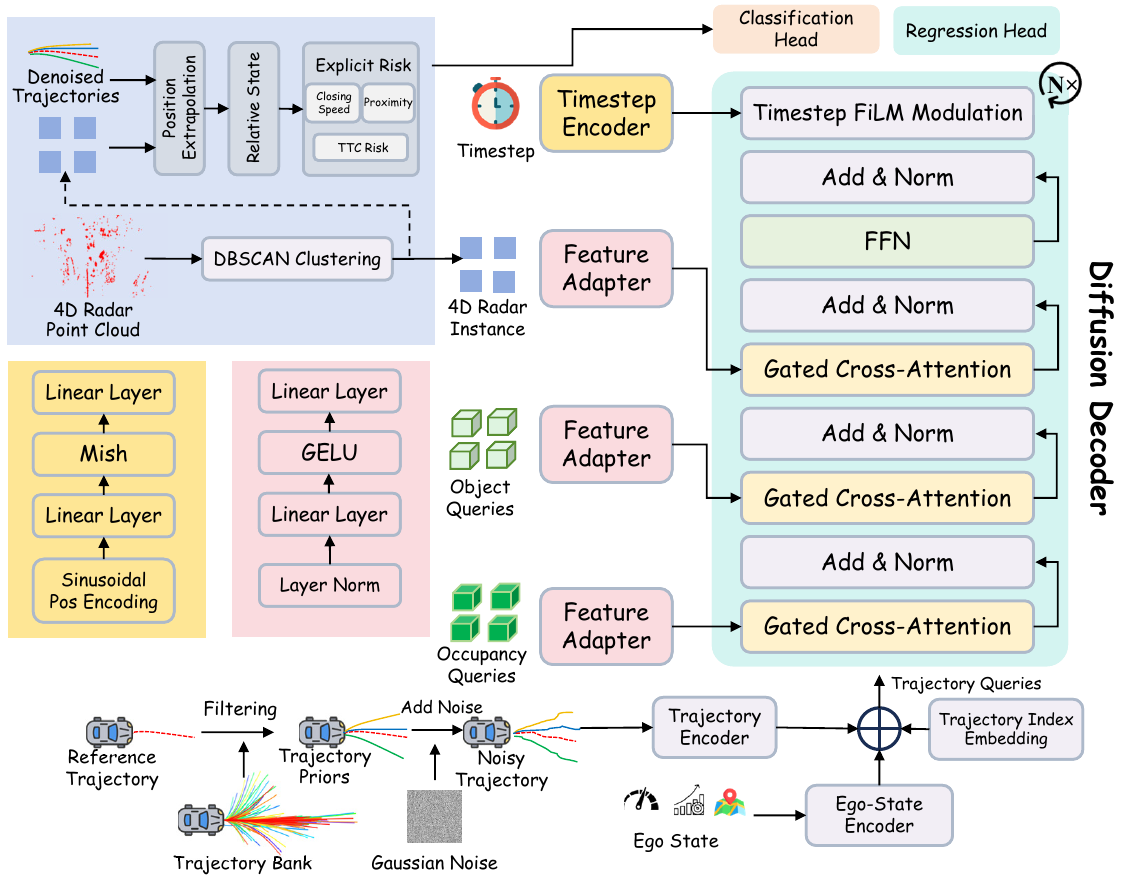}
  \caption{\textbf{Trajectory arbitration.}}
  \label{fig:physical_arbitration}
\end{wrapfigure}

As shown in Figure~\ref{fig:rcvla_overview}, RCVLA consists of two stages:
radar-grounded semantic reasoning (\textbf{RCVLA-Sem}) and
trajectory arbitration (\textbf{RCVLA-Phys}). In RCVLA-Sem,
multi-view image and 4D radar points are converted into fixed-length tokens
and enhanced through gated bidirectional interaction. The enhanced camera and radar tokens are fed into the LLM with the prompt and
ego features for driving question answering or long-horizon reference
planning, while the encoded sensor features produce object and occupancy
queries for subsequent trajectory arbitration. RCVLA-Phys combines the reference trajectory with selected anchors to form candidate priors, refines them through truncated diffusion conditioned
on structured scene queries and cluster-level radar measurements, and
calibrates their scores with radar-derived time-to-collision (TTC) risk for
final selection, as shown in Figure~\ref{fig:physical_arbitration}.

\subsection{Radar-Grounded Semantic Reasoning}
\label{sec:semantic_proposal}
\noindent\textbf{Modality tokens.}
We use EVA-02-L \citep{fang2024eva} to encode multi-view images into features
$\mathbf{F}_{\mathrm{I}}$, and RadarPillarNet \citep{zheng2023rcfusion} to extract 4D
radar BEV features $\mathbf{F}_{\mathrm{R}}$. The image features are organized
into a multi-view sequence with view and camera-aware 3D positional encodings,
while the radar BEV features are converted into a compact spatial sequence
with 2D positional encodings, yielding $\mathbf{M}_{\mathrm{I}}$ and
$\mathbf{M}_{\mathrm{R}}$. Following the query-based interfaces of BLIP-2 and
OmniDrive \citep{li2023blip,wang2025omnidrive}, independent IQ-Former and
RQ-Former modules use learnable queries $\mathbf{Q}_m^0$ to extract
fixed-length language-aligned tokens:
\begin{equation}
 \mathbf{T}_m=\operatorname{Proj}_m\!\left(
 \operatorname{QFormer}_m(\mathbf{Q}_m^0,\mathbf{M}_m)\right),
 \qquad m\in\{\mathrm{I},\mathrm{R}\}.
 \label{eq:qformer_tokens}
\end{equation}
where $\operatorname{QFormer}_m$ denotes IQ-Former or RQ-Former, and
$\operatorname{Proj}_m$ maps its query outputs to the LLM hidden space,
yielding $\mathbf{T}_{\mathrm{I}}$ and $\mathbf{T}_{\mathrm{R}}$.

\noindent\textbf{Gated bidirectional interaction.}
Direct interaction in the high-dimensional LLM space is computationally
expensive, while sparse and noisy radar observations can degrade cross-modal
interaction. We therefore project the language-aligned tokens into a compact
interaction space and apply token-wise gating to the radar features:
\begin{equation}
\mathbf{X}_m=
\operatorname{Down}_m\!\left(\operatorname{LN}(\mathbf{T}_m)\right).
\label{eq:interaction_projection}
\end{equation}
\begin{equation}
\mathbf{g}_{\mathrm R}
=
\sigma\!\left(
\operatorname{MLP}_{\mathrm g}
(\operatorname{LN}(\mathbf{X}_{\mathrm R}))
\right),
\qquad
\widehat{\mathbf{X}}_{\mathrm R}
=
\mathbf{g}_{\mathrm R}\odot\mathbf{X}_{\mathrm R}.
\label{eq:radar_gate}
\end{equation}
where $\operatorname{Down}_m(\cdot)$ denotes a modality-specific linear
projection, $\sigma(\cdot)$ is the sigmoid function, and
$\mathbf{g}_{\mathrm R}$ contains a scalar gate for each radar token.

The image branch attends to the gated radar features
$\widehat{\mathbf{X}}_{\mathrm R}$, while the radar branch attends to the
image features $\mathbf{X}_{\mathrm I}$. The two branches use separate
cross-attention modules, followed by residual and FFN updates, yielding
$\widetilde{\mathbf{X}}_{\mathrm I}$ and
$\widetilde{\mathbf{X}}_{\mathrm R}$. We retain the original language-aligned 
tokens and inject only the residual introduced by cross-modal interaction:
\begin{equation}
\mathbf{T}'_m
=
\mathbf{T}_m+
\alpha_m\operatorname{Up}_m
\left(\widetilde{\mathbf{X}}_m-\mathbf{X}_m\right).
\label{eq:delta_writeback}
\end{equation}
where $\operatorname{Up}_m(\cdot)$ projects the interaction residual back to
the LLM hidden space, and $\alpha_m$ is a learnable scalar initialized to a
small positive value.

\noindent\textbf{Language, reference trajectory, and structured queries.}
The enhanced image and radar tokens are combined with the prompt embeddings
and ego features and fed into a LoRA-adapted LLM \citep{hu2022lora}.
For QA tasks, the model generates language responses autoregressively. 
For planning prompts, we introduce a special token \texttt{<waypoint\_ego>} 
and pass its hidden state through an MLP to directly regress a continuous
long-horizon reference trajectory of 12 waypoints over the next
$6\,$s.

In parallel, two lightweight transformer decoders operate on the shared image
and radar features to produce object queries $\mathbf{Q}^{\mathrm{OD}}$ and
occupancy queries $\mathbf{Q}^{\mathrm{OCC}}$. The object branch uses
learnable 3D reference points, while the occupancy branch uses a regular 3D
grid. Both aggregate image and radar BEV features through deformable attention
and are supervised by 3D detection and semantic occupancy, respectively.
The resulting queries provide structured conditions for RCVLA-Phys.

\subsection{Trajectory Arbitration}
\label{sec:physical_arbitration}
\noindent\textbf{Reference-guided query initialization.}
We align the long-horizon reference trajectory from RCVLA-Sem to the current
ego frame and retain its next $3\,$s segment as a candidate prior. Additional
priors are retrieved from an offline trajectory-anchor bank to preserve both
reference consistency and motion diversity. Following truncated diffusion
\citep{liao2025diffusiondrive}, each candidate prior $\bm{\mu}_m$ is perturbed
as
\begin{equation}
 \mathbf{x}_m^i
 =\sqrt{\bar{\alpha}_i}\,\bm{\mu}_m
 +\sqrt{1-\bar{\alpha}_i}\,\bm{\epsilon}_m.
 \label{eq:candidate_noise}
\end{equation}
where $\bm{\epsilon}_m\sim\mathcal{N}(\mathbf{0},\mathbf{I})$ denotes
Gaussian noise and $\bar{\alpha}_i$ is the cumulative diffusion coefficient.
\noindent Each noisy prior is encoded by an MLP and combined with ego-state and
candidate-index embeddings to initialize the trajectory queries, which are
then passed to the diffusion decoder for iterative refinement.

\noindent\textbf{Structured diffusion decoder.}
\noindent The decoder refines trajectory queries using occupancy, object, and radar
conditions, which respectively provide free-space and obstacle constraints,
instance-level object representations, and direct measurements of position
and motion. Radar clusters are obtained from 4D radar points using DBSCAN \citep{schubert2017dbscan}. Feature adapters
project each condition into the trajectory-query feature dimension.

At decoder layer $\ell$, the trajectory queries interact with each condition
through cross-attention, and the attended context is incorporated through a
gated residual update:
\begin{align}
 \mathbf{C}_k^\ell
 &=\operatorname{MCA}_k\!\left(
   \operatorname{LN}(\mathbf{Q}_{k,\mathrm{in}}^\ell),\mathbf{M}_k,\mathbf{M}_k\right),
 \label{eq:conditional_attention}\\
 \mathbf{g}_k^\ell
 &=\sigma\!\left(\operatorname{MLP}_k
   [\mathbf{Q}_{k,\mathrm{in}}^\ell;\mathbf{C}_k^\ell]\right),
 \label{eq:conditional_gate}\\
 \mathbf{Q}_{k,\mathrm{out}}^\ell
 &=\operatorname{LN}\!\left(
   \mathbf{Q}_{k,\mathrm{in}}^\ell+
   \mathbf{g}_k^\ell\odot\mathbf{C}_k^\ell\right).
 \label{eq:conditional_update}
\end{align}
where $k\in\{\mathrm{OCC},\mathrm{OD},\mathrm{R}\}$ denotes the condition
type, $\mathbf{Q}_{k,\mathrm{in}}^\ell$ and
$\mathbf{Q}_{k,\mathrm{out}}^\ell$ are the trajectory queries before and after
interaction with condition $k$, and $\mathbf{M}_k$ contains the corresponding
adapted condition features. $\operatorname{MCA}_k(\cdot,\cdot,\cdot)$ denotes
multi-head cross-attention. The resulting condition context
$\mathbf{C}_k^\ell$ is modulated by the query-conditioned gate
$\mathbf{g}_k^\ell$ before being incorporated through
the residual update. The operators
$\sigma(\cdot)$ and $\odot$ denote the sigmoid function and elementwise
multiplication, respectively.
After the condition interactions, the trajectory queries are updated by an
FFN with a residual connection, yielding $\widetilde{\mathbf{Q}}^\ell$. The
diffusion timestep then modulates these queries through FiLM:
\begin{equation}
 \mathbf{Q}^\ell=\widetilde{\mathbf{Q}}^\ell\odot
 (\mathbf{1}+\bm{\gamma}_i^\ell)+\bm{\beta}_i^\ell,
 \label{eq:diffusion_film}
\end{equation}
where $\widetilde{\mathbf{Q}}^\ell$ denotes the trajectory queries after the
conditional interactions and FFN update, and $\bm{\gamma}_i^\ell$ and
$\bm{\beta}_i^\ell$ are the scale and bias generated from the
diffusion-timestep embedding. A regression head and a scoring head then
predict a denoised candidate trajectory $\widehat{\bm{\tau}}_m$ and its score
$s_m$, respectively.
The regression head directly predicts a coordinate correction to the noisy
trajectory rather than noise. At inference, the final-layer
decoder predictions update the candidates at the next DDIM timestep.

\noindent\textbf{Radar-based risk calibration.}
In addition to trajectory refinement, 4D radar clusters also provide physical
cues for risk-aware candidate selection. For radar cluster $j$, let
$\mathbf{p}_j$ and $\mathbf{v}_j$ denote its planar center position and
compensated radial-velocity vector, respectively. Assuming constant
velocity, its center at planning step $h$ is predicted as
$\mathbf{p}_{j,h}=\mathbf{p}_j+t_h\mathbf{v}_j$, where $t_h=h\Delta t$ and
$\Delta t=0.5\,\mathrm{s}$. For the $m$-th candidate
trajectory, let $\mathbf{e}_{m,h}$ and $\mathbf{v}_{m,h}$ denote the ego
position and velocity at step $h$, and let $d_{m,j,h}$ denote the minimum
planar distance between the ego and radar-cluster footprints, with
$d_{m,j,h}=0$ for overlap. The closing speed is
\begin{equation}
 c_{m,j,h}=\max\!\left(
 (\mathbf{v}_{m,h}-\mathbf{v}_j)^\top
 \frac{\mathbf{p}_{j,h}-\mathbf{e}_{m,h}}
 {\|\mathbf{p}_{j,h}-\mathbf{e}_{m,h}\|_2+\varepsilon},0\right).
 \label{eq:closing_speed}
\end{equation}
where $\varepsilon$ is a small constant for numerical stability. The
corresponding TTC is
\begin{equation}
 \mathrm{TTC}_{m,j,h}=\begin{cases}
 t_h,&d_{m,j,h}=0,\\
 t_h+d_{m,j,h}/c_{m,j,h},&d_{m,j,h}>0,\ c_{m,j,h}>0.
\end{cases}
\label{eq:waypoint_ttc}
\end{equation}
The minimum TTC over all clusters and waypoints is converted into candidate
risk:
\begin{equation}
 r_m=\begin{cases}
 \exp\!\left(-\mathrm{TTC}^{\min}_m/\tau_{\mathrm{TTC}}\right),
   &\mathrm{TTC}^{\min}_m\leq T_{\mathrm{h}},\\
 0,&\text{otherwise},
 \end{cases}
\label{eq:radar_risk}
\end{equation}
where $\mathrm{TTC}^{\min}_m=\min_{j,h}\mathrm{TTC}_{m,j,h}$ and
$T_{\mathrm{h}}$ is the planning horizon. A smaller TTC yields a larger risk.
During training, $r_m$ is used to construct a risk-aware soft ranking target,
while at inference it calibrates the candidate scores before final selection.

\subsection{Training Objectives}
\label{sec:training_objectives}
RCVLA is trained in three stages. First, the radar encoder is pretrained with
3D detection and semantic occupancy supervision. We then freeze the radar
encoder and the LLM and train only RQ-Former on Cap4DR with masked
autoregressive language modeling for radar--language alignment.

Second, we initialize the image and radar encoders, IQ-Former, and RQ-Former
from pretrained weights and fine-tune RCVLA-Sem on
OmniHD-QA. The base LLM remains frozen, while its LoRA adapters are
trainable. The objective jointly supervises language generation, reference
planning, and structured scene queries:
\begin{equation}
 \mathcal{L}_{\mathrm{sem}}=\mathcal{L}_{\mathrm{LM}}+\mathcal{L}_{\mathrm{wp}}
 +\lambda_{\mathrm{Det}}\mathcal{L}_{\mathrm{Det}}
 +\lambda_{\mathrm{OCC}}\mathcal{L}_{\mathrm{OCC}},
 \label{eq:semantic_loss}
\end{equation}
where $\mathcal{L}_{\mathrm{LM}}$ is the next-token prediction loss
and $\mathcal{L}_{\mathrm{wp}}$ is an $L_1$ loss. The detection and occupancy losses, $\mathcal{L}_{\mathrm{Det}}$ and
$\mathcal{L}_{\mathrm{OCC}}$, are defined following
\citep{zheng2026doracamom}.

Third, we freeze RCVLA-Sem and train RCVLA-Phys for trajectory refinement and
risk-aware candidate ranking. For $M$ denoised candidates, let
$\mathbf{d}$, $\mathbf{r}$, and $\mathbf{s}$ denote their average displacement
errors to the ground-truth trajectory $\bm{\tau}^{*}$, radar risk, and
predicted logits, respectively. The candidate with the smallest trajectory
error $m^{*}=\arg\min_m d_m$, is selected for regression. For candidate
ranking, we construct a risk-aware soft target $\bm{\pi}$ and calibrate the
predicted distribution $\mathbf{p}$ as
\begin{align}
\bm{\pi}
&=\operatorname{softmax}\!\left(
-\frac{\mathbf{d}+\lambda_{\mathrm r}\mathbf{r}}{T_{\mathrm c}}
\right),
\label{eq:risk_ranking_target}\\
\mathbf{p}
&=\operatorname{softmax}\!\left(
\mathbf{s}-\lambda_{\mathrm s}\mathbf{r}
\right),
\label{eq:risk_ranking}
\end{align}
where $\lambda_{\mathrm r}$ and $\lambda_{\mathrm s}$ control the contribution
of radar risk to the soft target and predicted logits, respectively, and
$T_{\mathrm c}$ is the temperature parameter. The overall objective jointly
optimizes trajectory regression and candidate ranking:
\begin{equation}
\mathcal{L}_{\mathrm{phys}}
=
\lambda_{\mathrm{reg}}
\operatorname{SmoothL1}\!\left(
\widehat{\bm{\tau}}_{m^{*}},\bm{\tau}^{*}
\right)
-
\sum_{m=1}^{M}\pi_m\log p_m,
\label{eq:physical_loss}
\end{equation}
where $\widehat{\bm{\tau}}_{m^{*}}$ is the selected denoised trajectory and
$\lambda_{\mathrm{reg}}$ balances the regression term. The regression candidate is determined solely by trajectory error, while radar risk affects only candidate ranking and score calibration.

\begin{wraptable}[21]{R}{0.58\textwidth}
\vspace{-5pt}
\centering
\caption{Scene description performance. B-$n$: BLEU-$n$; R-L: ROUGE-L;
MET.: METEOR. Higher is better.}
\label{tab:caption_results}
\rcvlatableformat
\scriptsize
\setlength{\tabcolsep}{0pt}
\renewcommand{\arraystretch}{0.96}
\begin{tabular*}{\linewidth}{@{\extracolsep{\fill}}lccccccc@{}}
\toprule
Method & B-1 & B-2 & B-3 & B-4 & R-L & CIDEr & MET. \\
\midrule
LLaVA-v1.5 & 14.50 & 8.08 & 4.90 & 2.91 & 19.53 & 7.25 & 19.31 \\
Qwen2.5-VL & 28.66 & 15.34 & 9.00 & 5.29 & 22.57 & 22.00 & 30.36 \\
Qwen3-VL & 30.53 & 17.31 & 10.74 & 6.78 & 24.35 & 24.38 & 33.57 \\
InternVL2.5 & 20.28 & 11.79 & 7.37 & 4.55 & 23.01 & 15.17 & 24.92 \\
InternVL3 & 26.85 & 14.82 & 8.82 & 5.28 & 23.05 & 20.45 & 28.28 \\
OmniDrive & 39.80 & 28.48 & 21.90 & 17.38 & 36.98 & 92.49 & 43.13 \\
\textbf{RCVLA-Sem} & \textbf{41.82} & \textbf{30.22} & \textbf{23.40} &
\textbf{18.69} & \textbf{38.28} & \textbf{102.41} & \textbf{44.91} \\
\bottomrule
\end{tabular*}
\centering
\caption{Occupancy and key-object reasoning performance. Occ., Dyn.,
Sta., and Drv. are class accuracies; $A_{\mathrm{occ}}$ and
$A_{\mathrm{int}}$ are aggregate accuracies. All values are percentages except
$V_{\mathrm{err}}$ (m/s).}
\label{tab:reasoning_results}
\rcvlatableformat
\scriptsize
\setlength{\tabcolsep}{0pt}
\renewcommand{\arraystretch}{0.96}
\begin{tabular*}{\linewidth}{@{\extracolsep{\fill}}lccccccc@{}}
\toprule
Method & Occ.$\uparrow$ & Dyn.$\uparrow$ & Sta.$\uparrow$ & Drv.$\uparrow$ &
$A_{\mathrm{occ}}\uparrow$ & $V_{\mathrm{err}}\downarrow$ &
$A_{\mathrm{int}}\uparrow$ \\
\midrule
LLaVA-v1.5 & 45.21 & 37.06 & 56.55 & 43.88 & 44.25 & 11.04 & 16.09 \\
Qwen2.5-VL & 47.89 & 25.48 & 56.56 & 15.17 & 35.00 & 6.16 & \textbf{75.63} \\
Qwen3-VL & 49.37 & 43.76 & 58.06 & 44.55 & 47.78 & 8.71 & 64.55 \\
InternVL2.5 & 56.30 & 45.31 & 45.24 & 90.46 & 61.51 & 5.83 & 75.22 \\
InternVL3 & 51.13 & 72.19 & 43.47 & 90.48 & 65.57 & 6.02 & 70.99 \\
OmniDrive & 70.30 & 82.44 & 69.44 & \textbf{92.15} & 78.81 & 3.61 & 75.38 \\
\textbf{RCVLA-Sem} & \textbf{72.93} & \textbf{82.57} & \textbf{75.29} & 91.67 &
\textbf{80.41} & \textbf{2.82} & 74.25 \\
\bottomrule
\end{tabular*}

\end{wraptable}
\section{Experiments}
\label{sec:experimental_setup}
\label{sec:experiments}

\textbf{Dataset and evaluation.}
Cap4DR is used only for RQ-Former pretraining, while all downstream
experiments are conducted on OmniHD-QA, with 358,185 training and 161,976
test samples. We evaluate scene description with standard caption metrics,
key-object reasoning with velocity error and intention accuracy, occupancy
reasoning with accuracy, and trajectory planning with open-loop L2 error and
collision rate. Baselines include
LLaVA-v1.5-7B \citep{liu2023visual},
Qwen2.5-VL-7B-Instruct \citep{bai2025qwen25vl},
Qwen3-VL-8B-Instruct \citep{bai2025qwen3},
InternVL2.5-8B \citep{chen2024internvl},
InternVL3-8B \citep{zhu2025internvl3},
OmniDrive \citep{wang2025omnidrive}, Constant-Velo,
BEV-Planner \citep{li2024ego}, and UniAD \citep{hu2023planning}.
The general VLM baselines perform zero-shot inference, while OmniDrive and UniAD are trained without map supervision.
All compared methods use the stated data split and evaluation metrics.
\textbf{Implementation details.}
We use six synchronized $640\times640$ camera views and multi-view 4D radar
points, with Vicuna-7B as the language model. IQ-Former and RQ-Former each contain 256 learnable queries and
six encoder layers. RCVLA-Phys uses five trajectory candidates, a two-layer
diffusion decoder, and two DDIM refinement steps. All stages are trained with
AdamW on four NVIDIA L20 GPUs.
The stated architecture and optimization configuration is used throughout.

\subsection{Main Results}
\label{sec:main_results}
\begin{wraptable}[21]{r}{0.58\textwidth}
\vspace{-10pt}
\centering
\caption{Open-loop planning on OmniHD-QA. L2 error and collision rate at each
horizon; lower is better.}
\label{tab:planning_results}
\scriptsize
\setlength{\tabcolsep}{0pt}
\renewcommand{\arraystretch}{0.96}
\begin{tabular*}{\linewidth}{@{\extracolsep{\fill}}lcccccccc@{}}
\toprule
& \multicolumn{4}{c}{L2 (m)} & \multicolumn{4}{c}{Collision (\%)} \\
\cmidrule(lr){2-5}\cmidrule(lr){6-9}
Method & 1 s & 2 s & 3 s & Avg. & 1 s & 2 s & 3 s & Avg. \\
\midrule
Qwen2.5-VL & 1.239 & 1.774 & 2.488 & 1.834 & 2.870 & 6.389 & 11.111 & 6.790 \\
Qwen3-VL & 0.224 & 0.581 & 1.059 & 0.621 & 0.370 & 2.623 & 6.420 & 3.138 \\
InternVL2.5 & 0.255 & 0.668 & 1.570 & 0.831 & 0.463 & 2.039 & 6.457 & 2.986 \\
InternVL3 & 2.411 & 4.646 & 7.086 & 4.714 & 19.136 & 23.796 & 31.636 & 24.856 \\
UniAD & 0.403 & 0.750 & 1.171 & 0.775 & 0.525 & 1.821 & 3.519 & 1.955 \\
Constant-Velo & 0.126 & 0.337 & 0.646 & 0.370 & 0.031 & 0.247 & 2.222 & 0.833 \\
BEV-Planner & 0.176 & 0.339 & 0.585 & 0.367 & 0.031 & \textbf{0.062} & 0.864 & 0.319 \\
OmniDrive & 0.121 & 0.334 & 0.648 & 0.368 & \textbf{0.000} & 0.123 & 1.420 & 0.514 \\
RCVLA-Sem & 0.117 & 0.317 & 0.608 & 0.348 & 0.031 & 0.216 & 1.481 & 0.576 \\
\textbf{RCVLA-Phys} & \textbf{0.109} & \textbf{0.236} & \textbf{0.431} &
\textbf{0.259} & \textbf{0.000} & 0.123 & \textbf{0.401} &
\textbf{0.175} \\
\bottomrule
\end{tabular*}
\centering
\caption{RCVLA-Sem ablation results. Accuracy and collision values are
percentages. RLP denotes radar-language pretraining.}
\label{tab:semantic_ablation}
\scriptsize
\setlength{\tabcolsep}{0pt}
\renewcommand{\arraystretch}{0.91}
\begin{tabular*}{\linewidth}{@{\extracolsep{\fill}}lcccccc@{}}
\toprule
Setting & CIDEr$\uparrow$ & $V_{\mathrm{err}}\downarrow$ &
$\mathrm{Acc}_{\mathrm{int}}\uparrow$ & $\mathrm{Acc}_{\mathrm{occ}}\uparrow$ &
L2$\downarrow$ & Coll.$\downarrow$ \\
\midrule
w/o 4D radar & 98.70 & 3.50 & 68.89 & 79.48 & \textbf{0.344} & \textbf{0.391} \\
w/o RLP & 101.60 & 3.45 & 70.97 & \textbf{80.63} & 0.345 & \textbf{0.391} \\
Token addition & 100.45 & 3.00 & 72.07 & 80.21 & 0.354 & 0.761 \\
Token concatenation & 102.20 & 2.96 & 73.80 & \textbf{80.63} & 0.353 & 0.525 \\
\textbf{Full} & \textbf{102.41} & \textbf{2.82} & \textbf{74.25} &
80.41 & 0.348 & 0.576 \\
\bottomrule
\end{tabular*}
\end{wraptable}

\begin{wrapfigure}[27]{r}{0.70\textwidth}
  \vspace{-2.6\baselineskip}
  \centering
  \includegraphics[width=\linewidth]{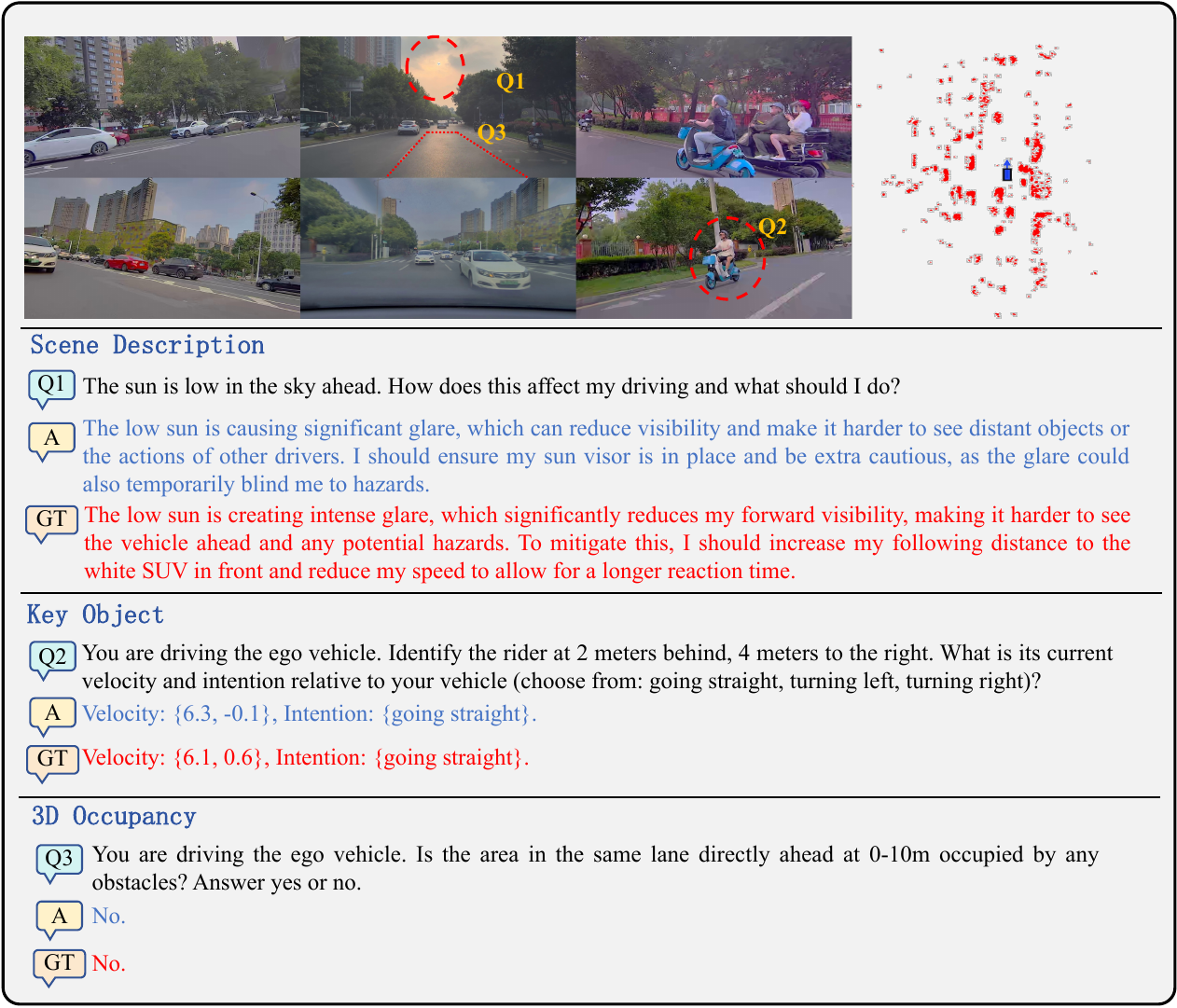}
  \captionsetup{skip=5pt}
  \caption{Qualitative question-answering results.}
  \label{fig:qa_visualization}
\end{wrapfigure}

\textbf{Scene cognition.}
RCVLA-Sem combines visual semantics with radar-derived spatial and
motion cues for scene description. In Table~\ref{tab:caption_results}, it
improves CIDEr over OmniDrive from $92.49$ to $102.41$, with gains across all
captioning metrics.
Table~\ref{tab:reasoning_results} shows that RCVLA-Sem improves
occupancy accuracy from $78.81\%$ to $80.41\%$ and reduces key-object velocity
error from $3.61$ to $2.82\,\mathrm{m/s}$. Drivable-area and intention
accuracies remain lower than OmniDrive by $0.48$ and $1.13$ points,
respectively.
Figure~\ref{fig:qa_visualization} provides qualitative question-answering
examples, illustrating how RCVLA-Sem combines visual semantics with
radar-derived spatial and motion cues for scene understanding.

\noindent\textbf{Trajectory planning.}
As shown in Table~\ref{tab:planning_results}, RCVLA-Phys achieves the lowest
average L2 error and collision rate among the compared methods, with
$0.259\,\mathrm{m}$ and $0.175\%$, respectively. Compared with RCVLA-Sem,
trajectory arbitration reduces the average L2 error from $0.348$ to
$0.259\,\mathrm{m}$ and the collision rate from $0.576\%$ to $0.175\%$.
Although RCVLA-Sem already outperforms BEV-Planner in average L2 error
($0.348$ vs.\ $0.367\,\mathrm{m}$), its collision rate remains higher
($0.576\%$ vs.\ $0.319\%$), indicating that accurate reference trajectories
do not necessarily correspond to lower collision risk. With RCVLA-Phys,
the L2 error is further reduced at all three horizons, while the $3\,\mathrm{s}$
collision rate drops from $1.481\%$ to $0.401\%$. Qualitative examples of
trajectory refinement are shown in Figure~\ref{fig:intro_trajectory_teaser}.

\subsection{Ablation Studies}
\label{sec:ablation}
\afterpage{%
\begin{figure}[t]
  \centering
  \begin{minipage}[t]{0.53\textwidth}
  \centering
  \includegraphics[width=\linewidth]{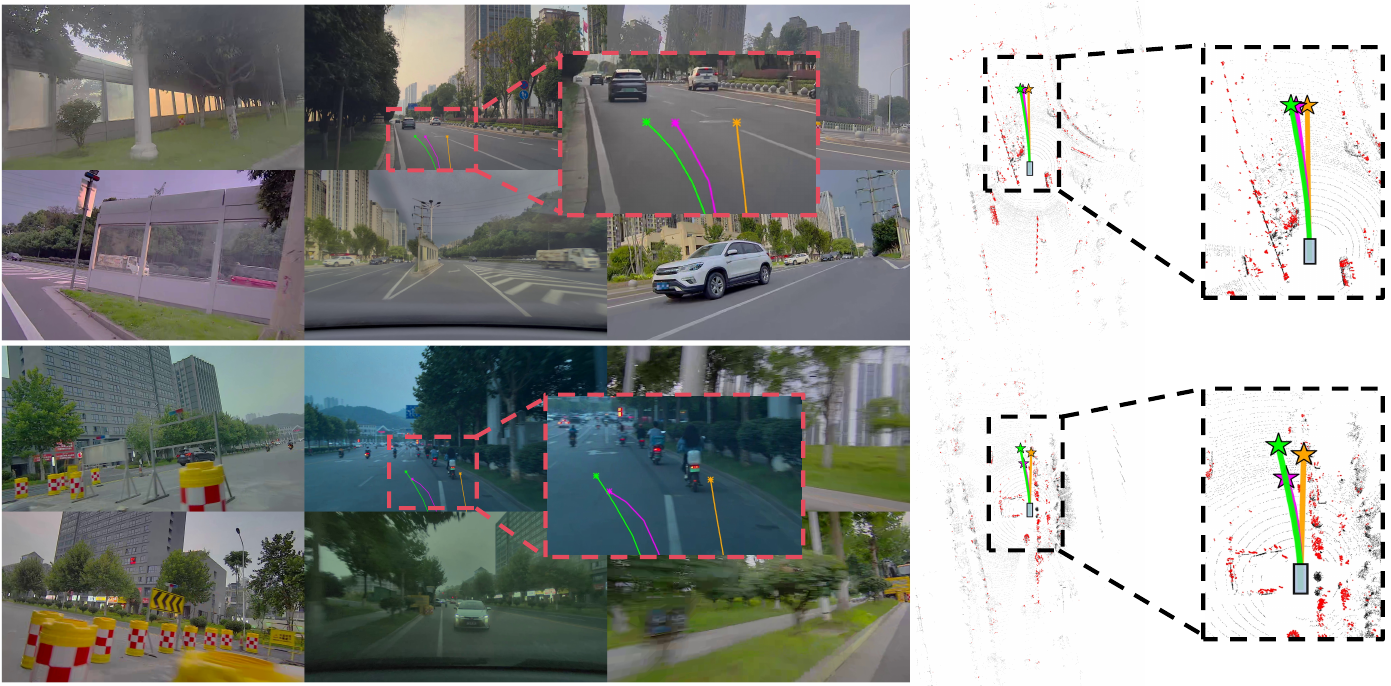}
  \captionof{figure}{Qualitative examples of trajectory arbitration. Orange,
magenta, and green denote the reference, arbitrated, and GT trajectories.}
  \label{fig:intro_trajectory_teaser}
  \end{minipage}\hfill
  \begin{minipage}[t]{0.46\textwidth}
  \centering
  \includegraphics[width=\linewidth]{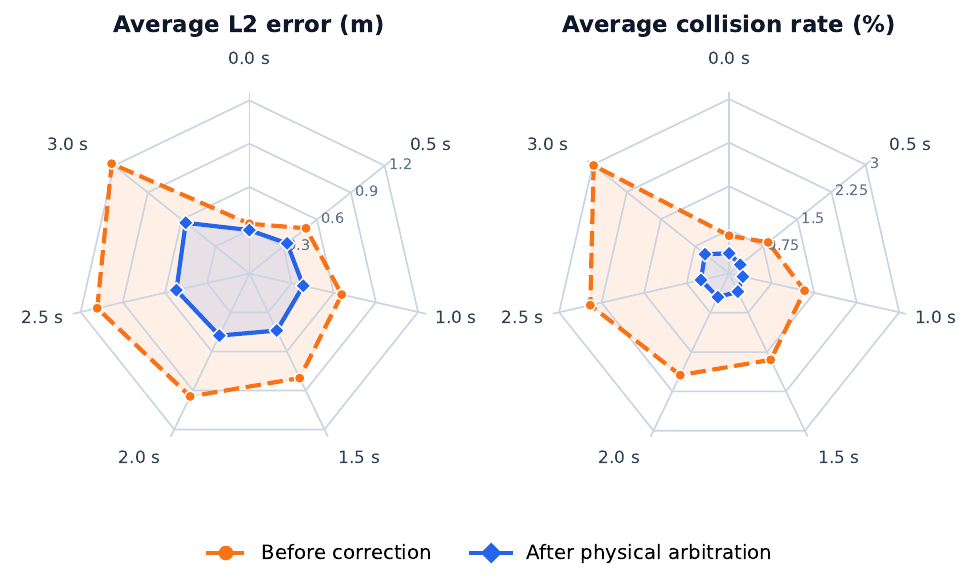}
  \captionof{figure}{Effect of reference delay on trajectory arbitration. Orange
  and blue denote the reanchored reference and arbitrated trajectory.}
  \label{fig:delay_curves}
  \end{minipage}
\end{figure}%
}
\textbf{RCVLA-Sem ablation.}
Table~\ref{tab:semantic_ablation} compares different RCVLA-Sem settings.
The full model achieves the best CIDEr, key-object velocity error, and
intention accuracy. Compared with removing 4D radar, it improves CIDEr from
$98.70$ to $102.41$ and reduces velocity error from $3.50$ to
$2.82\,\mathrm{m/s}$. Removing RLP or replacing the proposed interaction with
token addition or concatenation also degrades semantic performance. These
results support the roles of radar-language alignment and cross-modal
interaction in scene cognition.\WFclear
However, the gains in semantic reasoning do not translate into a lower
collision rate, suggesting that semantic-level radar integration alone is
insufficient for trajectory safety and motivating the dedicated trajectory
arbitration in RCVLA-Phys.

\begin{wraptable}[15]{r}{0.46\textwidth}
  \centering
  \captionof{table}{Ablation of RCVLA-Phys.}
  \label{tab:physical_ablation}
  \footnotesize
  \setlength{\tabcolsep}{0pt}
  \renewcommand{\arraystretch}{0.90}
  \begin{tabular*}{\linewidth}{@{\extracolsep{\fill}}lcc@{}}
    \toprule
    Setting & L2 (m)$\downarrow$ & Collision (\%)$\downarrow$ \\
    \midrule
    w/o occupancy queries & 0.321 & 0.597 \\
    w/o object queries & 0.320 & 0.309 \\
    w/o radar clusters & 0.297 & 0.442 \\
    w/o risk calibration & 0.296 & 0.340 \\
    \textbf{Full} & \textbf{0.259} & \textbf{0.175} \\
    \bottomrule
  \end{tabular*}

  \captionof{table}{Robustness to reference trajectories}
  \label{tab:reference_delay}
  \footnotesize
  \setlength{\tabcolsep}{0.4pt}
  \renewcommand{\arraystretch}{0.90}
  \begin{tabular*}{\linewidth}{@{\extracolsep{\fill}}lcc@{}}
  \toprule
  Delay & Avg. L2 (m)$\downarrow$ & Avg. Coll. (\%)$\downarrow$ \\
  \midrule
  0.0 s & 0.301 & 0.347 \\
  0.5 s & 0.335 & 0.243 \\
  1.0 s & 0.383 & 0.243 \\
  1.5 s & 0.439 & 0.347 \\
  2.0 s & 0.478 & 0.451 \\
  2.5 s & 0.519 & 0.498 \\
  3.0 s & 0.564 & 0.532 \\
  \bottomrule
  \end{tabular*}
\end{wraptable}

\textbf{RCVLA-Phys components.}
Table~\ref{tab:physical_ablation} evaluates the contribution of occupancy
queries, object queries, radar clusters, and radar risk calibration.
Removing occupancy queries causes the largest overall degradation, increasing
average L2 error from $0.259$ to $0.321\,\mathrm{m}$ and collision rate from
$0.175\%$ to $0.597\%$. Removing object queries increases the two metrics to
$0.320\,\mathrm{m}$ and $0.309\%$, respectively. Removing radar clusters
raises the collision rate to $0.442\%$, while removing risk calibration
increases it to $0.340\%$. These results show that all four components
contribute to trajectory arbitration, with both radar clusters and risk
calibration providing clear gains in collision reduction.

\label{sec:reference_perturbation}
\textbf{Robustness to delayed references.}
We evaluate RCVLA-Phys under delayed reference trajectories.
\WFclear
During training, reference delays are sampled from $0$ to $3\,\mathrm{s}$ at
$0.5\,\mathrm{s}$ intervals, while object, occupancy, and radar conditions
remain current. Expired waypoints are removed and the remaining trajectory is
reanchored to the current ego frame before arbitration.
As shown in Table~\ref{tab:reference_delay}, average L2 error increases from
$0.301$ to $0.564\,\mathrm{m}$ as the delay reaches $3\,\mathrm{s}$, while the
collision rate rises to $0.532\%$. Figure~\ref{fig:delay_curves} shows that
current scene conditions can partially compensate for delayed references, but
the degradation increases with reference staleness.

\WFclear
\FloatBarrier
\section{Conclusion}
\label{sec:conclusion}

In this work, we present RCVLA, a radar-camera VLA framework that exploits
4D radar in complementary roles for semantic reasoning and trajectory
arbitration. We construct Cap4DR for radar-language alignment and OmniHD-QA
for driving instruction tuning and evaluation. RCVLA-Sem performs
radar-grounded semantic reasoning through camera-radar token interaction,
supporting driving question answering and long-horizon reference planning
while producing structured object and occupancy queries. RCVLA-Phys further
refines reference-guided trajectory candidates using these queries and
cluster-level radar measurements, with radar-derived TTC risk calibrating
candidate selection. Experiments show that language-aligned radar tokens
improve scene cognition, while cluster-level radar measurements and risk
calibration further improve open-loop trajectory planning. Future work will
explore predictive world models for longer-horizon reasoning and planning.

\bibliography{iclr2027}
\bibliographystyle{iclr2027}
\end{document}